\documentclass[10pt,twocolumn,letterpaper]{article}

\usepackage[pagenumbers]{cvpr} 
\usepackage{graphicx}
\usepackage{amsmath}
\usepackage{amssymb}
\usepackage{booktabs}

\usepackage{makecell}
\usepackage{booktabs} 
\usepackage{diagbox}
\usepackage{multirow}
\usepackage{amssymb}
\usepackage{pifont}
\newcommand{\cmark}{\ding{51}}%
\newcommand{\xmark}{\ding{55}}%
\usepackage{float}
\usepackage [english]{babel}
\usepackage [autostyle, english = american]{csquotes}
\MakeOuterQuote{"}
\usepackage[normalem]{ulem}

\usepackage[pagebackref,breaklinks,colorlinks]{hyperref}

\usepackage[capitalize]{cleveref}
\crefname{section}{Sec.}{Secs.}
\Crefname{section}{Section}{Sections}
\Crefname{table}{Table}{Tables}
\crefname{table}{Tab.}{Tabs.}

\def\cvprPaperID{5413} 
\def\confName{CVPR}
\def\confYear{2022}

\begin{document}

\title{CoCAtt: A Cognitive-Conditioned Driver Attention Dataset}

\author{Yuan Shen\\
{\tt\small yshen47@illinois.edu}
\and
Niviru Wijayaratne*\\
{\tt\small nnw2@illinois.edu}
\and
Pranav Sriram*\\
{\tt\small psriram2@illinois.edu}
\and
Aamir Hasan\\
{\tt\small aamirh2@illinois.edu}
\and
Peter Du\\
{\tt\small peterdu2@illinois.edu}
\and
Katherine Driggs-Campbell\\
{\tt\small krdc@illinois.edu} \\
\and
University of Illinois at Urbana-Champaign
}

\maketitle
\begin{abstract}
The task of driver attention prediction has drawn considerable interest among researchers in robotics and the autonomous vehicle industry. Driver attention prediction can play an instrumental role in mitigating and preventing high-risk events, like collisions and casualties. However, existing driver attention prediction models neglect the distraction state and intention of the driver, which can significantly influence how they observe their surroundings. To address these issues, we present a new driver attention dataset, CoCAtt (Cognitive-Conditioned Attention). Unlike previous driver attention datasets, CoCAtt includes per-frame annotations that describe the distraction state and intention of the driver. In addition, the attention data in our dataset is captured in both manual and autopilot modes using eye-tracking devices of different resolutions. Our results demonstrate that incorporating the above two driver states into attention modeling can improve the performance of driver attention prediction. To the best of our knowledge, this work is the first to provide autopilot attention data. Furthermore, CoCAtt is currently the largest and the most diverse driver attention dataset in terms of autonomy levels, eye tracker resolutions, and driving scenarios.
\end{abstract}
\section{Introduction}

\begin{figure}[t]
\centerline{\includegraphics[width=0.5\textwidth]{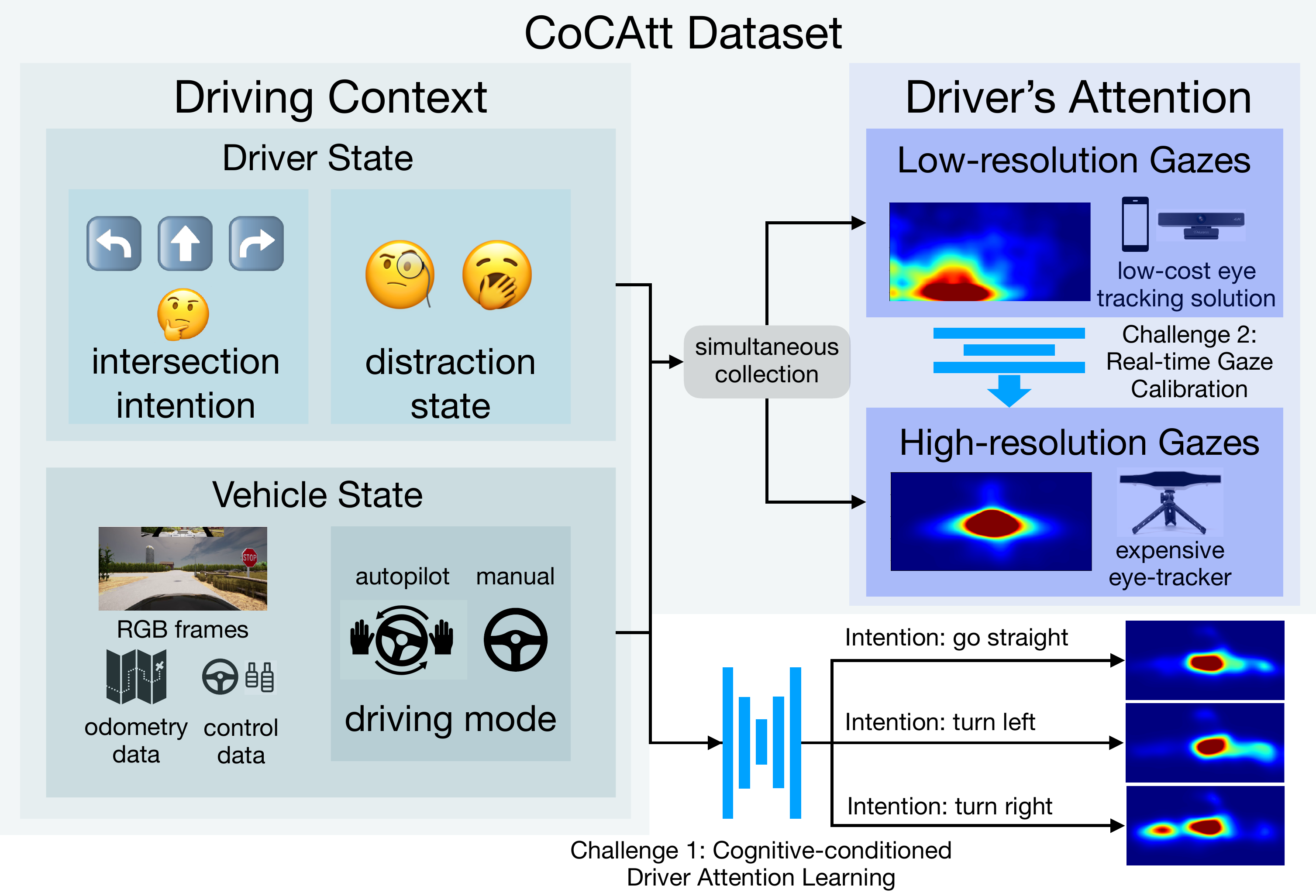}}
\caption{Our Driver Attention Dataset, CoCAtt, has the following novelties: (1) per-frame driver state annotations including distraction state and intersection intention (2) attention data in both manual and autopilot drive settings, and (3) gaze data simultaneously captured with both low and high-resolution eye-tracking devices.}
\label{fig:thumbnail}
\end{figure}

With recent breakthroughs in autonomous driving, the role of the human driver is evolving from performing all driving tasks to monitoring and assisting automation~\cite{janssen, rezvani2016towards, shia2014semiautonomous}.
However, as defined by SAE J3016 standards, human drivers still need to remain attentive at all times while driving at lower levels of autonomy, e.g., SAE Level 3~\cite{sae_level}. 
Thus, to better monitor driver behavior, researchers have shown great interest in modeling driver attention to capture the attention behavior of human drivers across driving scenes~\cite{BDDA, dreyeve, 9262067, s20072030, pal2020looking, baee2021medirl}.

The primary focus of the pioneering work of driver attention prediction has been in modeling generic driver attention behavior at a task-driven level, i.e., modeling what most drivers would attend to in the driving scene~\cite{dreyeve}.
For instance, under the popular in-lab driver attention collection protocol~\cite{BDDA}, the same driver-perspective video is shown to several participants to capture the generic attention of a driver in the driving scenarios presented in the videos.

However, despite their success in modeling the generic driver attention, existing methods tend to fail to accurately capture variations that arise from actual driving context.
Specifically, the state-of-the-art methods ignore two critical factors that affect driver attention.
First, current methods do not incorporate the internal state of the driver into their attention prediction frameworks. Internal state here is referred to the driver's distraction and intention.
Recent studies have found that the attention behavior of drivers is significantly influenced by their mental state~\cite{ojstervsek2019influence, magana2020effects}. 
It has also been discovered that task-driven factors and behavioral goals, i.e., intention, impact attention behavior~\cite{Hayhoe2005EyeMI, Hayhoe2014ModelingTC, Hayhoe2003VisualMA, Land2009LookingAA}.
Second, current methods do not incorporate the level of autonomy that the driver is operating under, which can also have a considerable effect on the attention behavior of the driver~\cite{Banks2018IsPA, 10.1145/3409120.3410644}. This dependence is primarily caused by the variance in cognitive load in maintaining situational awareness for human drivers with varying levels of autonomy.



In addition, the current driver attention modeling methods focus mostly on modeling the “expected” driver attention behavior but ignore the practical challenge of capturing the actual driver gaze in real-world vehicles. Specifically, many of the existing driver attention datasets use high-resolution but expensive eye trackers, e.g., the SMI ETG 2w Eye Tracking Glasses~\cite{smi_eye_tracker}, to record driver gaze~\cite{BDDA, dreyeve, DADA2000}. Although such equipment could improve driver monitoring, it is unlikely for drivers to wear or for manufacturers to install these expensive eye-tracking devices. Thus, a low-cost and accurate eye tracking solution is necessary but missing.

To address these issues, we propose a new driver attention dataset, CoCAtt (Cognitive-Conditioned Attention), with 11.88 hours of driver attention data, approximately twice the size of the current largest driver attention dataset, DADA-2000~\cite{DADA2000}. The main goal of our dataset is to collect driver attention data that captures both the cognitive factors and the task-driven nature of attention. Our dataset provides per-timestamp annotations for driver distraction state and intention, and provides gaze data in both autopilot and manual modes. Furthermore, to prepare training data for low-cost eye-tracking, we simultaneously collect attention data with eye-tracking devices in two different resolutions. In particular, to improve the quality of attention data captured with low-cost devices, we use data captured from a high-resolution eye tracker as ground truth. To demonstrate the utility of our dataset, we conduct experiments on our dataset with several baseline architectures. Our experiments suggest that incorporating additional information about the driver state can improve driver attention prediction. 

\section{Related Work} \label{sec:related_work}
\subsection{Driver Attention Datasets}
\label{sec:driver_attention_datasets}
 Four large-scale publicly available datasets have been popularly used in modeling driver attention: DR(eye)VE~\cite{dreyeve}, BDD-Attention~\cite{BDDA}, DADA-2000~\cite{DADA2000}, and EyeCar~\cite{baee2021medirl}. With 6 hours of eye-tracking data, the DR(eye)VE dataset is the only dataset collected in-car and also provides distraction-related annotations for 20\% of its frames~\cite{dreyeve}. However, there are two main limitations with this dataset. The first is that distraction-related annotations do not directly correspond to the actual internal state of the driver as those annotations are labeled during the post-processing steps by checking whether the gaze fixation is on driving-related objects. The second is that DR(eye)VE dataset has few interesting scenario with only one car per frame on average~\cite{BDDA}. The scenario issue is addressed by the Berkeley Deep-Drive Attention dataset (BDD-A), which contains many diverse driving scenarios in critical situations~\cite{BDDA}. 
 The attention data in the BDD-A dataset was collected in an in-lab setup, in which participants were asked to imagine themselves as the driver in a driver-perspective video. This collection protocol was later reused by the DADA-2000 and EyeCar project with a focus on traffic accident scenarios~\cite{DADA2000, baee2021medirl}. Although this collection protocol is cost-effective, the collected gaze data is less credible as the participants were not the actual drivers in the setup. Similar to our setup, the multi-modal dataset by Taamneh \etal, was also collected in a driving simulator under four different distraction conditions~\cite{taamneh}. However, with only a four-lane highway scenario, the driving scenario of Taamneh \etal is not as diverse as ours. In contrast, our data collection procedure allows participants to manually drive the car in an in-lab simulation environment, ensuring diverse driving scenarios and high gaze quality without sacrificing realism. For example in autopilot mode, we asked participants to put their hands on the wheel to resemble real-world driving. 
\begin{table*}[t]
\centering
\caption{Attribute comparison with existing driver attention datasets\label{table:table1}. The in-lab protocol refers to the method proposed by Xia \etal, in which gaze data was collected from participants who
were instructed to visualize themselves as the driver in driver-perspective videos~\cite{BDDA}.}
\renewcommand{\arraystretch}{1.0}  
 \begin{tabular}{@{}llllll@{}}
 \toprule
 Dataset & Duration & Collection Method & Field of View & Collection Type(s) & Scenario(s) \\ 
 \midrule
 DR(eye)VE~\cite{dreyeve} & 6.1 hrs & real-world driving & wide & manual & highway, countryside, downtown \\ 
 
 BDD-A~\cite{BDDA} & 3.5 hrs & in-lab protocol & narrow & N/A & downtown\\
 
 DADA-2000~\cite{DADA2000} & 6.1 hrs & in-lab protocol & N/A & N/A & accidents\\
 
EyeCar~\cite{baee2021medirl} & 3.5 hrs & in-lab protocol & wide & N/A & accidents\\

 \textbf{CoCAtt} & 11.9 hrs & simulator driving & wide & autopilot, manual & countryside\\ 
 \bottomrule
\end{tabular}
\end{table*}

\begin{table*}[t]
\centering
\caption{Data modality comparison with existing driver attention datasets.\label{table:table2}}
\renewcommand{\arraystretch}{1.0}
 \begin{tabular}{@{}lccccccc@{}}
 \toprule
 Dataset & \makecell{Front-View \\ Camera}& \makecell{Rear-View \\ Mirror} & \makecell{Distraction \\ Annotation} &\makecell{Intention \\ Annotation} & \makecell{Odometry \\ Sensor} & \makecell{Webcam \\ Gaze} & \makecell{Eye Tracker \\ Gaze} \\  
 \midrule
 DR(eye)VE~\cite{dreyeve} & yes & no  & \makecell{yes (post-hoc)} & no & yes & no & yes\\ 

 BDD-A~\cite{BDDA} & yes & no  & no & no & yes & no & yes\\

 DADA-2000~\cite{DADA2000} & yes & no & no & no & no & no & yes\\

 EyeCar~\cite{baee2021medirl} & yes & no & no & no & yes & no & yes\\

 \textbf{CoCAtt} & yes & yes  & yes & yes & yes & yes & yes\\ 
 \bottomrule
\end{tabular}
\end{table*}

\subsection{Driver Attention Models}

Early work on saliency map prediction proposed bottom-up, feature-based models which use low-level visual features such as color, intensity, and orientation in the captured scenes~\cite{ittietal, Koch1985ShiftsIS, hareletal, Torralba06contextualguidance, meuretal, erdem, liuetal}. For example, the pioneering work by Itti \etal proposed a framework for saliency-based visual attention in which feature maps are calculated using "center-surround" operations (accentuates locations that are distinct compared to their local surroundings) and are then normalized and integrated to produce a final saliency map. Another example is the Bayesian framework proposed by Torralba \etal, which integrates image saliency features calculated using a bottom-up computational model and contextual scene information to produce a saliency map. Other early methods also use machine learning models, e.g., SVMs, to predict saliency maps~\cite{bremond}. More recent methods predict driver attention using state-of-the-art deep learning techniques~\cite{baee2021medirl, dreyeve, pal2020looking}. 
For example, Alletto \etal proposed a 3D convolution-based multi-branch deep network to predict driver attention directly from raw video sequences using RGB image features, optical flow, and semantic segmentation~\cite{dreyeve}. 
Another example is the modular method proposed by Baee \etal that combines the front view and eye fixation data to extract context features for Deep Inverse Reinforcement learning~\cite{baee2021medirl}.  
There have also been efforts to improve models by combining gaze data with scene semantics so that only semantically relevant areas in the scene are considered~\cite{pal2020looking}. 

\subsection{Task-Driven Attention}
Bottom-up computational models rely solely on the visual features of the scene to guide attention, but it was later found that top-down factors, such as task and behavioral goals, are also linked to attention~\cite{Hayhoe2005EyeMI, Hayhoe2014ModelingTC, Hayhoe2003VisualMA, Land2009LookingAA}. The seminal works by Hayhoe and Ballard describe the relationship between a subject's cognitive processes and their gaze fixations and how these fixations are highly dependent on the task itself. They found that fixations are geared towards obtaining the most valuable task-specific information and thus the subject may not necessarily fixate on the most salient regions in a scene~\cite{Hayhoe2005EyeMI, Hayhoe2014ModelingTC}. The effect of task on visual attention has given rise to task-driven attention models, which have shown to be effective for visual attention prediction~\cite{gao2019goaloriented, amadori}. For example, Gao \etal present the new problem of Object Importance Estimation, and propose a framework composed of a visual model and a goal model to directly incorporate the effect of driving goal into the task at hand. Another example is the architecture proposed by Amadori \etal named HammerDrive, in which attention maps are predicted separately for three defined maneuvers  (left lane change, right lane change, lane keep) and are then aggregated together into a final attention map weighted by a feature vector generated based on the vehicle maneuver. 

\section{The CoCAtt Dataset} \label{sec:dataset}
We present a cognitive-conditioned driver attention dataset, CoCAtt, collected in the CARLA simulation environment~\cite{Dosovitskiy17}. Our dataset contains 11.88 hours of driving data, twice the total duration of the current largest driver attention dataset, DADA-2000~\cite{DADA2000}. We provide a comparison of the statistics between our dataset and previous driver attention datasets in Tables~\ref{table:table1} and~\ref{table:table2}. The detailed statistics of our dataset are listed in Table~\ref{tab:table:Lumos_attention}. We highlight the following novelties of our dataset:
\begin{enumerate}
\setlength\itemsep{0em}
    \item CoCAtt is the first cognitive-conditioned driver attention dataset that provides per-frame annotations of driver state, including driver distraction and intention.
    \item CoCAtt is the first driver attention dataset that contains driver gaze data in both automated driving (SAE Level 3) and manual driving settings.
    \item CoCAtt provides gaze data captured simultaneously with eye-tracking devices of varying resolution. 
\end{enumerate}
The first contribution aims to provide a robust dataset that enables driver attention models to incorporate cognitive-conditioned learning into their framework. The second contribution aims to explore the effects of autonomy level on driver attention. Improved modeling of driver attention when using mid-levels of automation can help further research in human-vehicle collaboration and driver monitoring~\cite{rahman, Wang2021HumanVehicleCO, jha,rezvani2017optimizing}.
Finally, the third contribution aims to resolve the eye-tracking dilemma between cost and precision. Our insight is that a learning-based solution can calibrate low-resolution gaze by implicitly matching driving scene information with gaze fixations. Gathering eye-tracking data of varying resolution can be treated as a first step towards high precision eye tracking with low-cost, third-party devices, e.g., mobile phones.   

\subsection{Participants}
Our dataset contains samples from 11 participants, aged between 19 and 28 years old, who have at least one year of driving experience. Due to limitations of eye-tracking devices, we restricted participants to those who do not wear glasses while driving. 
We consider this to be a limitation on the hardware side that can be potentially improved in the future by eye-tracker manufacturers~\cite{9304573}.

\subsection{Hardware and Software Setup}
\begin{figure}
\centerline{\includegraphics[width=0.5\textwidth]{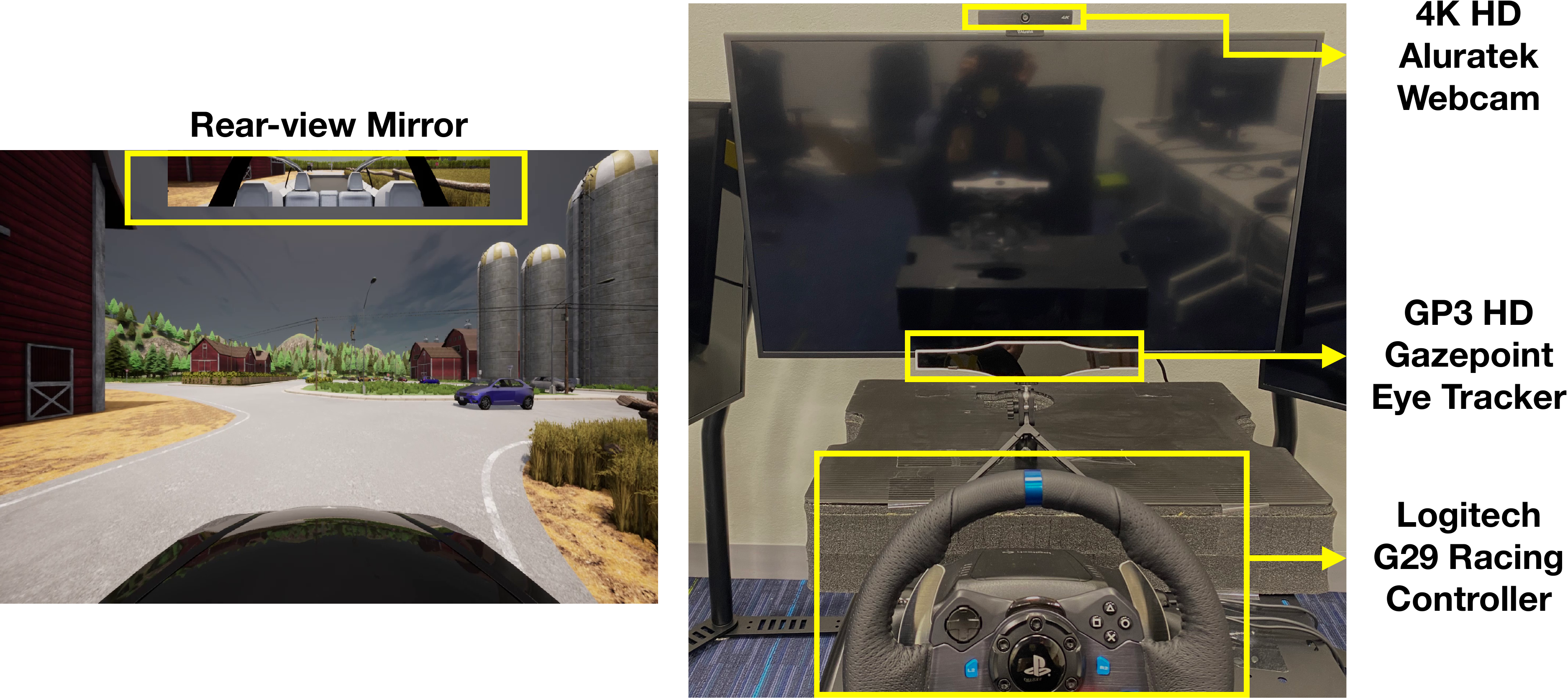}}
\caption{Data collection software interface and hardware setup.}
\label{fig:driving_simulator_interface}
\end{figure}

The participants drove a Vesaro car simulator equipped with a Logitech G29 wheel controller as shown in Figure~\ref{fig:driving_simulator_interface}. We used a Gazepoint GP3 eye tracker with 60 Hz and a webcam with 4K resolution at 30 Hz to collect participants' attention data. 
We used GazeRecorder, an online webcam eye-tracking platform, to capture webcam gaze (low-resolution gaze), and Gazepoint Analysis for the eye-tracker capture (high-resolution gaze). To simulate the driving scenarios, we set up our driving environment in the CARLA simulator as shown in Figure \ref{fig:driving_simulator_interface}. 
To include diverse driving scenarios, we used the CARLA default map, Town07, which covers stop-sign, yield-sign, and traffic light intersections. At the start of each session, we spawned 50 vehicles and 20 walkers at random locations. To ensure that participants could see oncoming vehicles at intersections, we set the field of view of the main camera to be 130 degrees. For automated driving, we used the roaming agent provided by CARLA 0.9.11.

\subsection{Distraction Task}

\begin{table*}[h]
\centering
    \caption{CoCAtt Dataset Statistics.}
    \label{tab:table:Lumos_attention}
    \renewcommand{\arraystretch}{1.0}
    \begin{tabular}{@{}lccccccccccc@{}} 
    \toprule 
      \multirow{2}{*}{Mode} & \multirow{2}{*}{\makecell{\small{Total} \\ \small{Sessions}}} & \multicolumn{3}{ c}{Intersection Types} & \multicolumn{3}{ c}{Intersection Actions} & \multicolumn{2}{ c}{\makecell{Distraction States}} & \multirow{2}{*}{\makecell{\small{Eye Tracker}}} & \multirow{2}{*}{\small{Webcam}} \\
      \cmidrule(lr){3-5} \cmidrule(lr){6-8} \cmidrule(lr){9-10}
      & & \small{Yield} & \small{Stop} & \small{Traffic Light} & \small{Left} & \small{Right} & \small{Forward} & \small{Distracted} & \small{Attentive} & & \\
      \midrule
      autopilot & 26 & 277 & 306 & 158 & 235 & 222 & 284 & 2.37 hrs & 2.87 hrs & yes & no \\
      manual & 42 & 372 & 590 & 260 & 304 & 409 & 509 & 2.83 hrs& 3.81 hrs & yes & yes \\
      \bottomrule
    \end{tabular}
\end{table*}


To obtain the ground-truth labels for the driver's cognitive state, we randomly prompted verbal distraction tasks to control each participant's distraction state during the driving sessions. For the distraction task, we conducted a pilot study to decide between two options: the n-back test~\cite{owen2005n} and the auditory clock angle test used in \cite{7378954}. Despite its popularity in cognitive science literature, our pilot study participants considered the n-back test too overwhelming to handle during driving and not representative of the actual distracted mental activities that drivers engage in. On the other hand, the clock angle test is closer to the distracted scenario, e.g., deciding where to eat by imagining the visual appearance of restaurants and foods. Thus, we chose the clock angle test in our study. We distributed distraction tasks in 20-second intervals throughout each drive session. For each distraction interval, three randomly generated timestamps, e.g., "10:30 AM", were prompted by a text-to-speech engine, with 4-second gaps between each timestamp to allow the participant to respond. Following each prompted timestamp, participants verbally answered whether the clock hands of an analog clock at that time form an acute angle. 

\subsection{Data Collection Procedure}

The duration of data collection for a single participant was two hours. To allow the participant to become comfortable with the CARLA simulation environment, we allowed a 20-minute warm-up session. The participants were first asked to drive without any prompts to familiarize themselves with the controls. They were then told to respond to randomly distributed distraction tasks to verify their understanding of the tasks was correct. In addition, they were asked to follow instructions when approaching oncoming intersections, e.g., "turn left". Our instructions aimed to control the driver's intention during manual driving sessions and ensure full map coverage. 

During the autopilot sessions, we simulated the automation level for SAE Level 3, in which the driver is required to maintain situational awareness throughout the entire session in case of an unplanned transfer of control~\cite{rezvani2016towards}. While observing surroundings, participants responded to the distraction tasks. Similar to the collection protocols of DADA-2000 and BDD-A, participants could not take over control of the car, even though we asked them to be prepared for emergency take-over~\cite{DADA2000,BDDA}. Furthermore, participants had no preview of the planned action of the ego vehicle. However, our collection procedure is unique in that participants had access to both tactile and visual stimuli. We find this multi-modal experience to be more realistic for participants and a more effective data collection procedure than the in-lab collection protocols discussed in Section \ref{sec:driver_attention_datasets}~\cite{multimodal_benefits}. During the manual drive sessions, the participants drove the ego vehicle while performing the distraction tasks. They were instructed to obey traffic rules and not to overtake any cars. As previously mentioned, the participants were verbally informed in advance about which directions to drive for incoming intersections. To avoid drowsiness, we broke down all sessions into 10 to 15-minute sub-sessions with 3 to 5-minute breaks in between. 

During the post-processing steps, similar to the Dr(eye)ve project, we aggregated the gaze fixations over nearby past and future frames, 0.01 second away, to the current timestamp. We applied Gaussian filter to smooth the eye fixation. Additionally, we filtered out blink and saccade eye movements.

\subsection{Dataset Analysis}
\label{sec:dataset_analysis}

\begin{figure}[t]
\centerline{\includegraphics[width=\columnwidth]{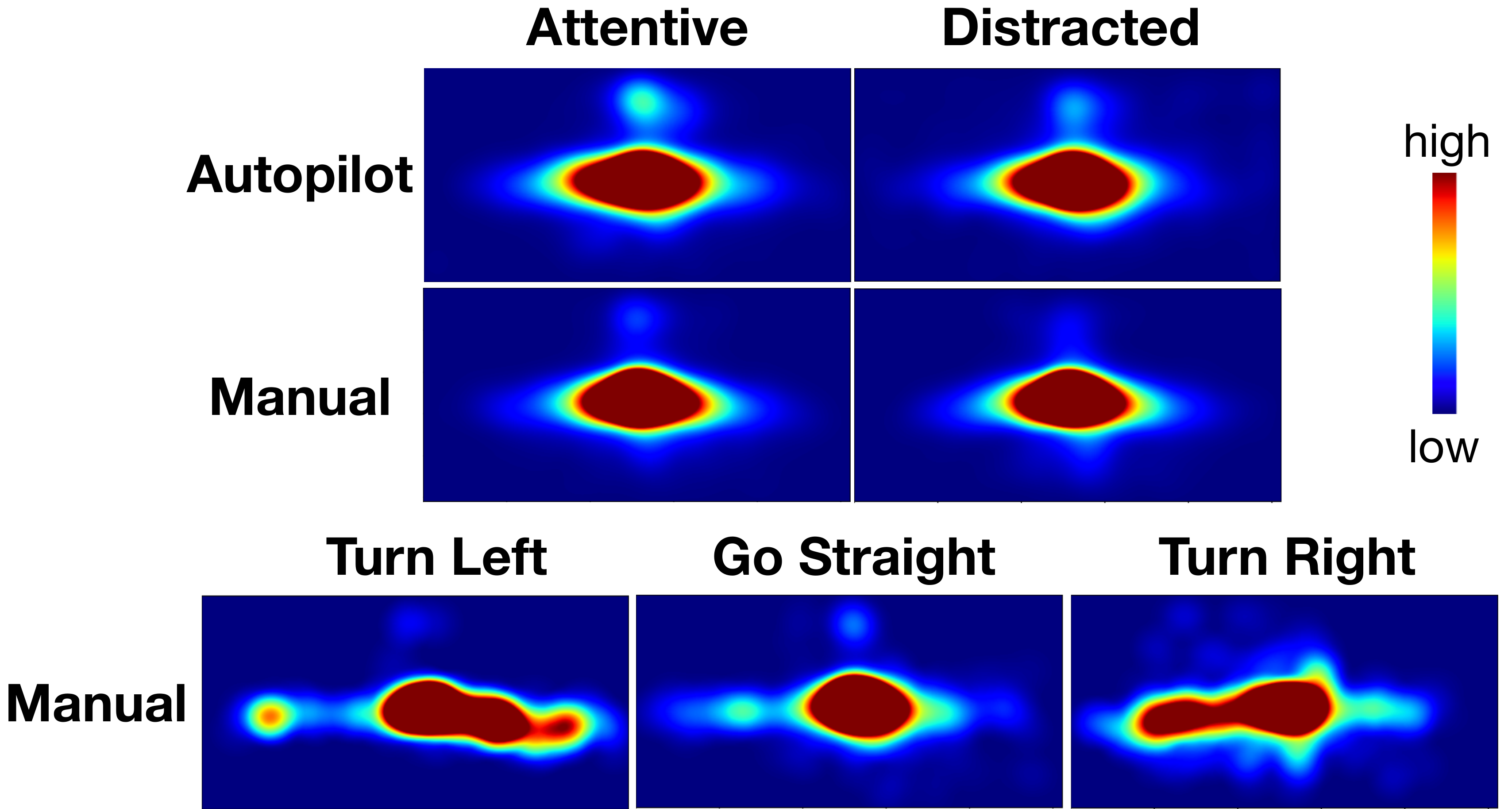}}
\caption{\label{fig:cumulative_heatmap} Cumulative heat map visualization of driver attention under different driver states. Distraction~(\emph{top}): under both distracted and attentive conditions, we uniformly sample the frames of attention maps based on location and average over those maps in road segments that are 30 meters away from intersections.  Intention~(\emph{bottom}): similarly, we aggregate attention maps captured when the ego vehicle is 30 meters away and proceeding towards any four-way intersection. We exclude traffic-light intersections during aggregation due to the dependence of attention on traffic light signals. We clip the normalized maps between 0 and 0.05 for all the above cumulative heatmaps for visualization purposes. }
\label{fig:cumulative-heatmap}
\end{figure}

To validate the need for driver state in attention prediction, we analyze the collected data to determine how much the driver intention and distraction influence attention. Moreover, we explore how the attention distribution varies across the autopilot and manual settings. With that, we sum up sequences of gaze maps with the same drive setting and cognitive (intention or distraction) and visualize the normalized attention heatmaps in Figure~\ref{fig:cumulative_heatmap}. 

Our results show several interesting findings. First, driver intention can influence their attention, especially towards the boundary region of the driving scene. For instance, the driver is more likely to check for oncoming cars from the right when turning left and vice-versa. Second, in terms of the effect of distraction state, although its impact on attention is weaker than that of intention, we find that attention during distraction shows stronger tunnel vision tendencies than during attentive states with more weights allocated at the center of the driving scene and less on the rear mirror region. Third, in terms of the effect of driver modes, we discover that the influence of distraction on driver attention is stronger in autopilot mode than manual drive mode. We also observe that attention in the rear mirror region is a critical feature that can differentiate drivers in two different driving modes and different driver states. Specifically, an attentive driver is more likely to check the rear mirror region than a distracted driver. Participants in autopilot mode checked the rear mirror regions more frequently than in the manual drive setting. 

\section{CoCAtt Open Challenges\label{sec:challenges}}
\subsection{Cognitive-conditioned Attention Learning}
Driver attention depends on both the external environment and human's internal states, e.g., their distraction state. Without consideration of driver internal state, attention modeling can be inaccurate. To incorporate the internal state of the driver into attention prediction, we formulate the following cognitive-conditioned attention model:
\begin{equation*}
    \hat \theta = \text{argmax}_\theta P_\theta(\hat A_t|c_{1:t}, I_{1:t})
\end{equation*}

\begin{figure*}[t]
\centerline{\includegraphics[width=0.75\textwidth]{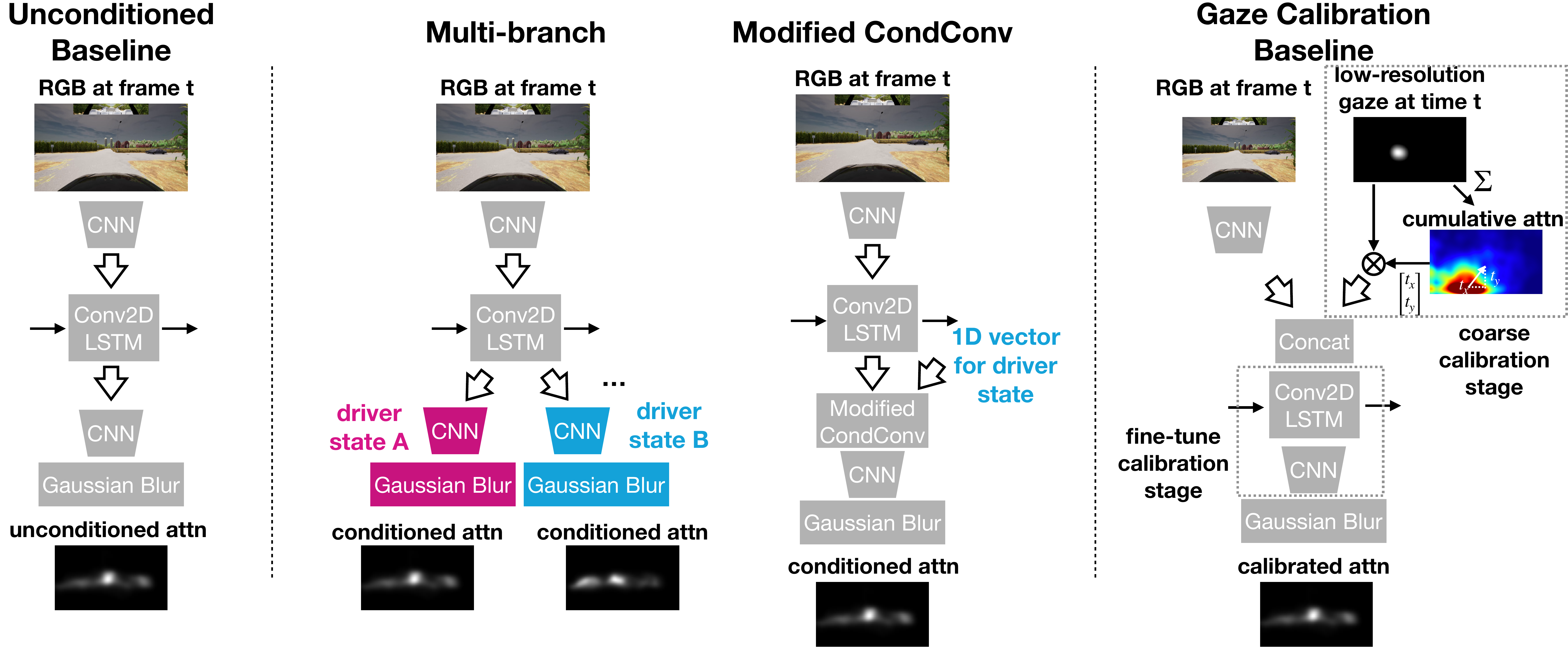}}
\caption{Our baseline architectures for the two proposed CoCAtt challenges. We use the architecture of BDD-A as our unconditioned baseline. For the task of cognitive-conditioned attention learning, we explore two baseline architectures, i.e., multi-branch and modified CondConv. For the former case, different network sub-branches are used to model different driver states, while the latter relies on conditional convolution to incorporate driver states. As for the gaze calibration baseline, we introduce a two-scale calibration procedure, in which we calculate a coarse shift offset based on previous attention maps and then fine-tune through a recurrent convolution network.}
\label{fig:baseline_model}
\end{figure*}
where $\hat A$ is the predicted 2D attention map, $c$ are driver states, $I$ are RGB frames, and $\theta$ are model parameters. In this work, we consider two types of driver states, including intersection intention and distraction state. In practice, intersection intention can be extracted from navigation applications, e.g., Google Maps, for each intersection, and distraction state can be inferred from user behavior data, e.g., eye blink frequency and pupil diameter~\cite{hybrid_bn, 4220657}.

\subsection{Real-time Gaze Calibration}
Existing work rely heavily on high-resolution eye-trackers for driver attention-related applications to monitor drivers in real-time or collect driver attention data for training~\cite{BDDA, dreyeve, DADA2000}. However, the high cost of those eye trackers reduces scalability, as it is difficult to manufacture and deploy high-cost eye trackers at a large scale. With eye-tracking data collected using two types of eye-tracking devices in CoCAtt, we propose the task of real-time gaze calibration over the attention data collected with low-cost eye-tracking devices, e.g., webcam or phone cameras. Formally,
\begin{equation*}
    \hat \theta = \text{argmax}_\theta P_\theta(\hat A_t^h|A^l_{1:t}, I_{1:t})
\end{equation*}
where $A^l_t$ represents low-resolution attention map at time t captured with low-cost eye-trackers and $\hat A_t^h$ denotes the calibrated high-resolution attention map at time t. In this work, we use attention data captured with GP3 eye tracker data as the ground-truth $A_t^h$, and attention data captured with 4K Aluratek webcam as $A_t^l$.

\section{Baseline Architectures \label{sec:baseline_architecture}}
As illustrated in Figure \ref{fig:baseline_model}, we introduce several baseline architectures for the two proposed tasks. The main purpose of these baseline architectures is to demonstrate the feasibility and potential of the proposed tasks. With that being said, we do not claim any architecture design related contribution and leave model improvement for future work. The implementation details are included in the supplementary material.
\subsection{Cognitive-Conditioned Attention Learning}
One key consideration in using the additional driver state information provided in CoCAtt is how it can be incorporated into the existing attention network. We explore two different architecture designs, i.e., multi-branch and modified CondConv, which can be easily adapted to existing state-of-the-art driver attention models. For the multi-branch architecture, we have a shared backbone to extract spatial-temporal visual features, and then use different branches to predict attention for different driver states. The advantage of the multi-branch architecture is that branches for new driver state conditions can be easily added to the pre-trained models without affecting the performance of other driver state conditions. However, a disadvantage of this architecture is that the number of attention heads increases linearly with the number of driver states, making model size a potential bottleneck to deployment on small devices in practice. Thus, we explored another light-weight solution via a CondConv layer to incorporate driver state through input-dependent convolution kernels~\cite{NEURIPS2019_f2201f51}. Note that the original CondConv only takes feature maps from the previous layer as inputs, however, we modified the CondConv by taking driver state as the input to the routing functions. The details of the modified CondConv are included in our supplementary materials. Compared to the multi-branch architecture, a shortcoming of this approach is that adding additional driver state conditions can potentially affect performance on other driver states.

\subsection{Real-time Gaze Calibration}
To demonstrate the feasibility of high-resolution eye-tracking using low-cost devices, we propose a simple but effective baseline architecture that uses a two step coarse-to-fine calibration step. In the coarse calibration component, to address the issue of noisy webcam gaze data~(check supplementary material for details), we first aggregate past low-resolution gaze maps over a fixed-size sliding window and calculate a coarse shift offset. The coarse shift offset is calculated based on the distance between the center coordinate in the scene and the position of the highest density in the aggregated gaze map. To further fine-tune the attention, a recurrent convolution network calibrates the centered gaze maps by implicitly matching scene feature points with the coarse gaze map.
\section{Experiments}\label{sec:experiment}
\begin{table*}[t]
  \centering
    \caption{Quantitative comparison between cognitive-conditioned and unconditioned attention learning\label{table:quantitative_res_condition}}
    \renewcommand{\arraystretch}{1.0}
    \begin{tabular}{@{}lcllllll@{}} 
    \toprule
      \multirow{2}{*}{Model} & \multirow{2}{*}{\makecell{Cognitive \\ Conditioned}} & \multicolumn{3}{c}{Intersection Intention}  & \multicolumn{3}{c}{Distraction State} \\
       \cmidrule(lr){3-5} \cmidrule(lr){6-8}
        & & $CC\uparrow$ & $H\downarrow$ & $D_{KL}\downarrow$ & $CC\uparrow$ & $H\downarrow$ & $D_{KL}\downarrow$\\
      \midrule
      unconditioned & no & \textbf{0.48} & 5.69 & 1.60 &
      0.45 & 5.70 & 1.58
      \\
      multi-branch baseline & yes &
      0.47 & \textbf{5.51} & 1.64 &
      0.45 & 5.63 & 1.55 
      \\
      modified CondConv & no &
      0.47 & 5.55 & 1.58 &
      \textbf{0.46} & 5.75 & \textbf{1.50} 
      \\
      modified CondConv & yes &
      \textbf{0.48} & \textbf{5.51} & \textbf{1.56} &
      \textbf{0.46} & \textbf{5.59} & \textbf{1.50} 
      \\
      \bottomrule
    \end{tabular}
\end{table*}
We conduct rigorous experiments with the baseline models on our CoCAtt dataset for both of the proposed tasks.
For the task of cognitive-conditioned attention, we start by verifying whether incorporating driver state information can improve the performance of driver attention prediction in Section \ref{sec:exp2}. We then explore the generative capabilities of the cognitive-conditioned model. For the gaze calibration task, we demonstrate the potential of leveraging low-resolution eye-tracking data to improve driver attention prediction with the proposed baselines in Section~\ref{sec:exp3}. In addition to the main experiments above, we also report the investigations of the effect of driving modes on driver attention modeling. Finally, to demonstrate the practicality of the cognitive-conditioned attention model, we showcase its use case on road safety analysis. Both the latter experiment and the use case are included in the supplementary material. 
\subsection{Experimental Setup}
\textbf{Evaluation Metrics}:  Since the ground-truth saliency map in our dataset is continuous, we use two distribution-based metrics for both tasks: Pearson Correlation Coefficient~($CC$) and KL-divergence~($D_{KL}$). $D_{KL}$, as an asymmetric dissimilarity metric, which penalizes false negatives more than false positives. On the other hand, $CC$ weighs both false negatives and false positives equally. Additionally, we report prediction entropy~($H$) for the saliency map to quantify prediction confidence. Our definition of entropy is included in the supplementary material.
For real-time gaze calibration, since we simultaneously record participants' attention data with two types of eye-tracking devices, we use eye-tracker attention data as the ground truth.

\textbf{Dataset Preparation}:
 For the task of cognitive-conditioned attention, to control the interaction effect, we split CoCAtt into two subsets for intention and distraction-state conditioned training. For the intention data, we sample frame sequences when the ego car approaches an intersection but has not yet performed any action at the intersection. We define each sequence to begin 30 meters away from the intersection and end when the ego car is closest to the intersection center. Note that, we only include stop-sign intersections for training. For the distraction-related experiments, we use frames pertaining to lane-following scenarios, i.e., segments that are 30 meters away from intersection. Since we do not inform the participant about the intersection action of the ego-car beforehand in the autopilot mode, we only use manual-drive data for the intention-conditioned case. For simplicity, we also only use autopilot-mode data for the distraction-conditioned case. To ensure equal consideration for different intentions, we prepare the validation and test sets by extracting 20 sequences for each intention per set while ensuring no overlaps. We apply re-weighted sampling based on the sequence number of each condition type (intention or distraction state) for all our experiments to ensure balanced sample during training. As for the task of real-time gaze calibration, we use all of the data collected in autopilot mode in which attention data is captured with both types of eye-tracking devices. The video frames are sampled at 4 fps with a sliding window size of 64. The sampled image resolutions are 256x512, and the ground-truth gaze map resolution for eye tracker and webcam eye-tracking data is 32x64.

\textbf{Implementation Details}:
We used AlexNet with ImageNet pre-trained weights to extract visual features similar to BDD-A. We use the cross-entropy loss to train all of the baseline models. We used Adam optimizer with weight decay~(learning rate at 0.0001, $\beta_1$ = 0.9, $\beta_2$ = 0.999, $\epsilon$ = 1e-8). Our experiments were trained on Tesla V100 GPU until convergence with a batch size of 9. The batch size here is selected to maximize GPU memory usage during training.

\subsection{Experiments On the Effects of Driver States}\label{sec:exp2} 
To verify whether knowledge of driver states can improve attention learning, we conduct a comparison study by training our model with and without the driver state. 

Our quantitative results suggest that cognitive-conditioned learning can improve driver attention prediction. As demonstrated in Table~\ref{table:quantitative_res_condition}, our quantitative results indicate that cognitive-conditioned attention shows less noisy predictions with reduced entropy in both types of driver states. And all metrics are improved for the modified CondConv baseline with driver state conditioning. 

\begin{figure}[ht]
\centerline{\includegraphics[width=0.42\textwidth]{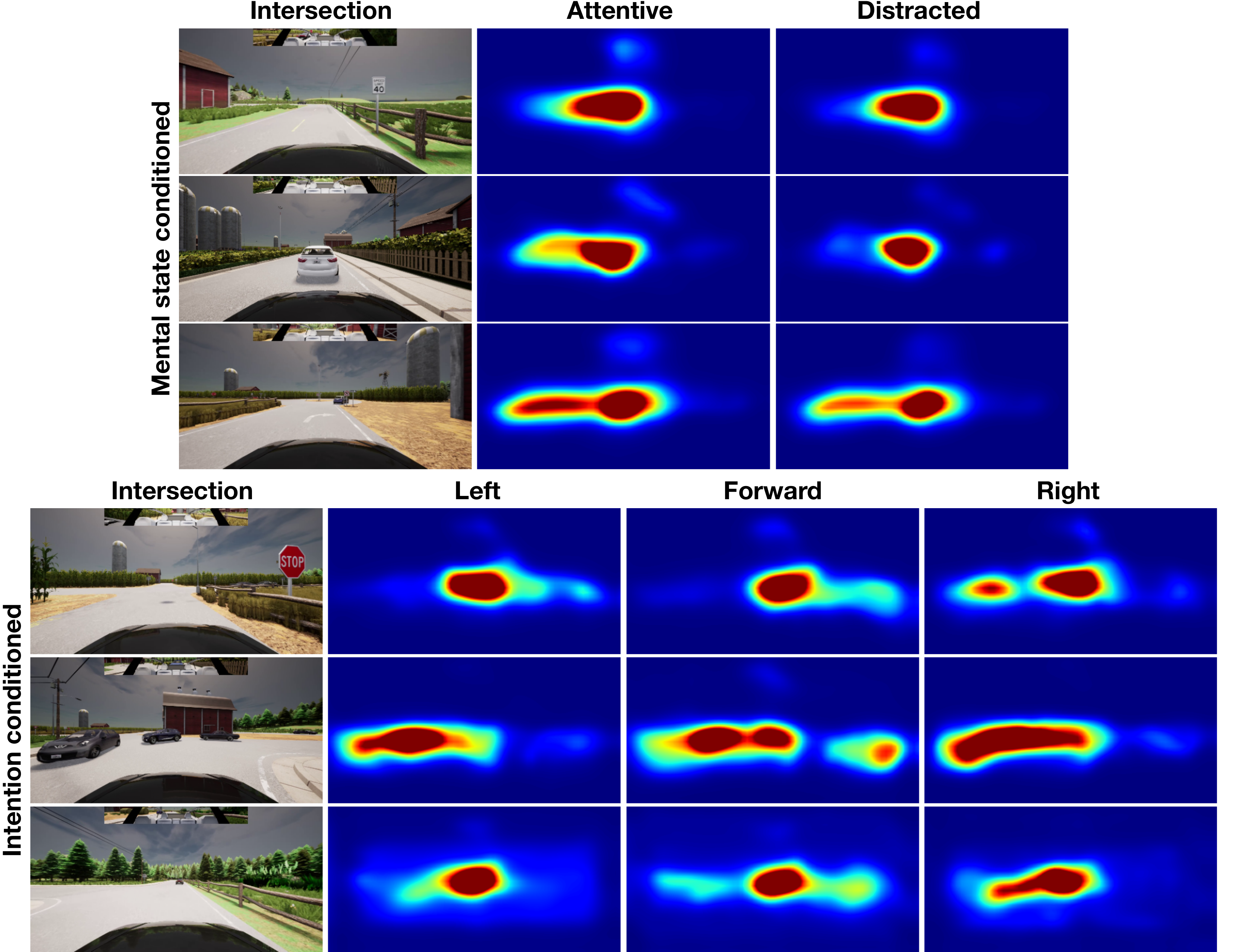}}
\caption{Qualitative results for two types of conditioned attention predictions from the multi-branch baseline. Each heatmap corresponds to the predicted cumulative attention if the driver state was set with the respective column condition. \label{fig:exp2_qualitative_whatif_res}}
\end{figure}

 To more deeply explore the conditional generative quality of our cognitive-conditioned model, we vary the conditions (intention and distraction state) while keeping other scene-related data untouched. As illustrated in Figure \ref{fig:exp2_qualitative_whatif_res}, our generative results under both driver state conditions show attention predictions that align with our analysis on human attention behaviors in Section \ref{sec:dataset_analysis}. For example, all the distracted cases show a stronger tendency to focus on the scene center than the attentive cases. For the intention-conditioned result, shown in the first row, the generative attention at intersections also shows a strong tendency, similar to human drivers, to observe the oncoming car from the opposite direction when turning. However, we notice that our generative model can give incorrect predictions for unrealistic scenarios not present in the training set. As shown in the last row of the intention-conditioned case, despite no road on the right, the predicted attention shows attention prediction, looking for cars from the left. In reality, drivers would be more likely to check on the right since there is no road. 
 
 Aside from the qualitative study, we conducted a human evaluation to measure whether cognitive-conditioned attention matches human belief better than unconditioned attention. All of our 11 participants have at least one year of driving experience. Our evaluation is conducted through a Google survey form with 18 binary choice questions. For each question, based on an intersection scene image and a pre-specified intersection action, the participants need to select the cumulative attention map that better matches their belief about the driver's attention at that situation. As shown in Table \ref{table:quantitative_generative_exp}, participants agreed more with the conditioned attention prediction for all three of the intersection intentions. 

\subsection{Experiments On the Effects of Gaze Calibration}
\label{sec:exp3}
Table~\ref{table:exp3_ablation_study} shows our ablation study results. Our quantitative results illustrate the importance of two-scale calibration procedures, as the centered and fine-tuned webcam gaze achieved higher performance than the other baselines. To further verify that webcam input plays a more significant role in the calibrated output than driving scene features, we provide a qualitative video study in supplementary material.


\begin{table}[t]
  \begin{center}
    \caption{Our human evaluation results on the cognitive-conditioned attention models. Each column lists the percentage of total scenarios with the specified intention that participants selected as matching their potential gaze. \label{table:quantitative_generative_exp}}
    
    \renewcommand{\arraystretch}{1.0}
    \begin{tabular}{@{}ccllll@{}} 
    \toprule 
      \multirow{2}{*}{Model} & \multicolumn{3}{c}{Intersection Intention} & \multirow{2}{*}{Total}\\
      \cmidrule(lr){2-4}
       & Left & Right & Forward & \\
       \midrule
       unconditioned & 44$\%$ & 32$\%$ & 40$\%$ & 39$\%$ \\ 
       conditioned & 56$\%$  & 68$\%$ & 60$\%$ & 61$\%$ \\ 
    \bottomrule
    \end{tabular} 
  \end{center}
\end{table}

\begin{table}[t]
  \begin{center}
    \caption{Ablation study for two-scale gaze calibration. Results are compared against collected eye-tracker data. The Coarse column indicates whether the coarse calibration centering operation was applied. The Fine-tune column indicates whether upstream webcam gaze was fed into the recurrent convolution network. \label{table:exp3_ablation_study}}
    
    \renewcommand{\arraystretch}{1.0}
    \begin{tabular}{@{}ccllll@{}} 
    \toprule 
      Coarse & Fine-tune & $D_{KL}\downarrow$ & $CC\uparrow$ & $H\downarrow$\\
      \midrule
       \xmark &\xmark & 10.83 & 0.14 & 18.66 \\ 
       \cmark & \xmark  & 8.61 & 0.25 & 19.62 \\ 
     \xmark & \cmark & 1.64 & 0.50 & 5.63 \\ 
      \cmark & \cmark & \textbf{1.57} & \textbf{0.52} & \textbf{5.44} \\ 
      \bottomrule
    \end{tabular} 
  \end{center}
\end{table}
\section{Limitations}

There are several limitations of our work. First, during the data collection process for autopilot mode, we do not inform the participants explicitly about the ego-car actions for upcoming intersections, and thus we cannot directly compare autopilot and manual drive attention around intersection regions. Second, when we collect the low-resolution attention data with a webcam, we do not introduce any noise caused by camera movements, which can frequently occur in real-world driving. Third, the data is collected in simulation environment, which might not match exactly with real-world driving. Finally, all of our baseline models for cognitive-conditioned attention assume driver state is known. An end-to-end system that can jointly infer attention and driver states is required in practice. 


\section{Conclusion}\label{sec:conclusion}
Our main contribution is a large-scale driver attention dataset, which aims to fill in the data gap of missing driver internal states in the existing driver attention dataset. Our work also seeks to draw the community's attention to the distinctive driver attention behavior in autonomous driving mode to better monitor human attention in future. Finally, we highlight the conflict of high-resolution eye-tracking and expensive costs in practice. To promote low-cost eye-tracking, we provide one potential solution of leveraging scene information to calibrate gaze. However, our current work does not provide a one-stop solution for the above issues, and thus further exploration is necessary in future.

\section{Acknowledgement}
This work was supported by State Farm and the Illinois Center for Autonomy. This work utilizes resources supported by the National Science Foundation’s Major Research Instrumentation program, grant \#1725729, as well as the University of Illinois at Urbana-Champaign~\cite{10.1145/3311790.3396649}. We thank Zhe Huang for feedback on paper drafts.
{\small
\bibliographystyle{ieee_fullname}
\bibliography{egbib}
}

\end{document}